\documentclass[journal ]{new-aiaa}
\usepackage[utf8]{inputenc}
\usepackage{textcomp}

\usepackage{appendix}
\usepackage{graphicx}
\usepackage{amsmath}
\usepackage[version=4]{mhchem}
\usepackage{siunitx}
\usepackage{longtable,tabularx}
\usepackage{multicol}
\usepackage{subfigure}
\usepackage{blindtext}
\usepackage{enumitem}
\usepackage{rotating}
\usepackage{pdflscape}
\usepackage{adjustbox}
\usepackage{mathtools}
\usepackage{pdfpages}

\title{Artificial Neural Network Modeling for Airline Disruption Management}

\author{Kolawole Ogunsina \footnote{Ph.D. Candidate, Purdue University, 701 W Stadium Avenue, West Lafayette, IN 47907, AIAA Student Member}}
\affil{School of Aeronautics and Astronautics, Purdue University, West Lafayette, IN 47907, United States.}
\author{Wendy A. Okolo \footnote{Aerospace Research Engineer, NASA Ames Research Center, Moffett Field, CA 94035, AIAA Member}}
\affil{Intelligent Systems Division, NASA Ames Research Center, Moffett Field, CA 94035, United States.}

\begin{document}

\maketitle

\begin{abstract}
Since the 1970s, most airlines have incorporated computerized support for managing disruptions during flight schedule execution. However, existing platforms for airline disruption management (ADM) employ monolithic system design methods that rely on the creation of specific rules and requirements through explicit optimization routines, before a system that meets the specifications is designed. Thus, current platforms for ADM are unable to readily accommodate additional system complexities resulting from the introduction of new capabilities, such as the introduction of unmanned aerial systems (UAS), operations and infrastructure, to the system. To this end, we use historical data on airline scheduling and operations recovery to develop a system of artificial neural networks (ANNs), which describe a predictive transfer function model (PTFM) for promptly estimating the recovery impact of disruption resolutions at separate phases of flight schedule execution during ADM. Furthermore, we provide a modular approach for assessing and executing the PTFM by employing a parallel ensemble method to develop generative routines that amalgamate the system of ANNs. Our modular approach ensures that current industry standards for tardiness in flight schedule execution during ADM are satisfied, while accurately estimating appropriate time-based performance metrics for the separate phases of flight schedule execution. 
\end{abstract}




\section{Introduction}
\lettrine{A}{irline} disruption management (ADM) can be categorized into tactical disruption management (tactical ADM), operational disruption management (operational ADM), and strategic disruption management (strategic ADM) \citep{Ogunsina2021a}. Tactical ADM defines proactive airline scheduling prior to the execution of a flight schedule on day of operation, operational ADM defines airline rescheduling during the execution of a flight schedule on day of operation, while strategic ADM represents proactive airline scheduling after the execution of a flight schedule. Many models for the processes that affect different phases of flight (and ADM) are defined by explicit objective functions, decision variables, and constraints, through optimization methods that impose a set of airline business rules from aircraft boarding at the origin airport to aircraft gate-parking at the destination airport \citep{Barnhart2003}. These optimization methods place a restriction on the search rules used to obtain a feasible solution for a flight schedule feature (i.e. decision variable), thereby enabling a parochial routine for estimating said features during ADM. Some of these models provide decent estimates of different flight schedule features for operational ADM during schedule execution \citep{Petersen2012, Marla2017, Lee2018}. However, their solution processes require considerable amount of time to evaluate the objectives and constraints posed by unscalable local and global routines that define the optimization of a monolithic system.

Current research and practices on ADM have primarily focused on minimizing the propagation of flight delays in the air transportation network during operational (reactive) ADM, by attempting to predict flight delay and its effect on flight schedule management. Many of these practices heuristically study rudimentary mechanisms (i.e. rules) of operation in the air transportation network to assess delay duration for airline schedule recovery and disruption management \citep{Beatty1999, Zhang2008, Fricke2009, Balakrishna2010}. Moreover, the recent emergence of data-driven methods enable the estimation of flight delay directly through data mining, in lieu of exploring existing flight delay propagation mechanisms \citep{Ye2020}. As such, the status quo for evaluating the performance of disruption resolutions during ADM relies on precision and accuracy in estimating flight delay duration. While flight delay duration provides a credible performance metric (or proxy) for assessing reactive ADM during schedule execution, it can not be used to readily evaluate proactive ADM initiatives because these initiatives are applied at different times prior to schedule execution. Furthermore, proactive ADM initiatives are typically addressed and employed by different departments within an airline other than the airline operations control center (AOCC). These departments spend significant time and resources in scheduling flight time and turnaround duration during proactive airline scheduling prior to flight schedule execution. This is done to alleviate flight delays during schedule execution on day of operation \citep{Hao2013}. Thus, there is a need to ensure an integrated solution for ADM by utilizing appropriate time-based performance metrics for different phases of ADM. 

Senior management in most AOCCs have historically argued that the measure of the performance of a corrective action for resolving disruption is only viable for reactive ADM during schedule execution. Thus, there are several industry-wide metrics that human specialists in the AOCC use to examine the quality of their corrective actions for disrupted flight schedules during irregular operations. Some of these metrics include \citep{Amadeus2016}:
\begin{itemize}
    \item Re-accommodation time period required for all passengers on a disrupted flight schedule.
    \item Time period required to create a plan of delays (or cancellations) that will enable the resumption of a disrupted flight schedule.
    \item The capacity for a disrupted flight schedule to depart exactly as scheduled, and arrive on schedule or within $14 mins$ of original schedule.
    \item The amount of delayed or cancelled flight legs in a flight schedule.
\end{itemize}
Although each of the aforementioned metrics is ultimately relevant for improving the quality of ADM solutions, each one only addresses an aspect of mitigation approaches for irregular operations during airline schedule execution. To enable an objective, all-inclusive and integrated methodology that measures the performance of irregular operations for a disrupted flight schedule, a candidate intelligent agent for ADM must be capable of expediently estimating the performance of alternative disruption resolutions (or flight schedules) by applying multiple performance metrics. Supervised learning techniques provide a suitable medium for creating models that can promptly estimate useful target features to appraise the performance of disruption resolution initiatives at different phases of ADM. Thus, we propose a predictive transfer function model (PTFM) to mine historical data on flight schedules and their corresponding performance on airline resource management. This will allow for a calibration of functional structures (i.e., find and apply new search rules) that represent efficient tools for an intelligent specialist agent in the AOCC. These tools enable prompt and precise estimation of appropriate measures of performance for disruption resolutions (i.e. alternate flight schedules) during tactical, operational, and strategic phases of ADM.  
\subsection{Contributions}
We adopt supervised machine learning methods (i.e. classification and regression) to characterize and evaluate multiple functional roles (i.e. domains) in the AOCC of a major U.S. airline. To achieve the aforementioned objectives, this paper enhances prior research on ADM through the following contributions:
\begin{enumerate}
    \item We expertly use historical data on airline scheduling and operations recovery from a major U.S. airline to develop a system of artificial neural networks that defines a predictive transfer function model (PTFM) framework. The PTFM framework estimates different measures of time for separate phases of ADM.
    \item We provide a modular approach to assess and execute the PTFM, by employing a parallel ensemble method to develop generative routines that amalgamate the system of artificial neural networks. 
\end{enumerate}

\subsection{Paper Organization}
Section~\ref{PTFM_method} discusses the role of the artificial neural network in the definition of the PTFM and the methodology used to formulate and assemble the PTFM. Next, Section~\ref{setup} discusses the computational setup of the PTFM for a representative functional role in the AOCC, followed by Section~\ref{results} that discusses the analyses and results from the application of the PTFM. Section \ref{conclusion} concludes with a summary of pertinent findings and areas for further research.

\section{The Predictive Transfer Function Model}\label{PTFM_method}
\begin{figure}[h!]
	\centering
	\includegraphics[width=0.99\textwidth]{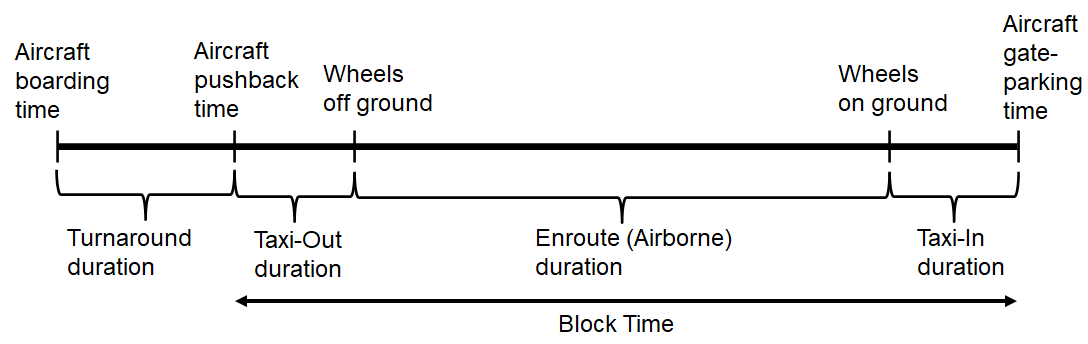}
	\caption{Timeline horizon during flight schedule execution}
	\label{fig:ADM_FTD}
\end{figure}

From a passenger's perspective, evaluation of the quality of service provided by an airline typically starts from aircraft boarding at a departure station's gate and ends upon aircraft parking at the arrival station's gate \citep{Midkiff2004}. As such, airlines aim to ensure optimal efficiency in the operational processes that ensue during the events in the flight schedule execution horizon shown in Fig.~\ref{fig:ADM_FTD}.

There are two primary stages during the execution of a characteristic flight schedule on day of operation, based upon an airline's resource capacity to serve its passengers. The first stage, known as \textit{flight capacity management}, defines the airline's ability to effectively load and unload an aircraft with passengers and necessary flight provisions (such as food and drinks), while intermittently conducting minor aircraft servicing tasks. Thus, from an airline operations perspective, the main objective of flight capacity management is to precisely estimate the time period required to complete the aircraft boarding (and deplaning) process during flight schedule execution, which represents the turnaround duration shown in Fig.~\ref{fig:ADM_FTD}. As such, a secondary objective of flight capacity management is to minimize unnecessary holdup (i.e. tactical delay) during the turnaround process.
 
The second stage of flight schedule execution also known as \textit{flight service management} defines the capacity of the airline and other air transportation stakeholders (such as air traffic control) to efficiently move the aircraft (loaded with passengers) from a designated gate at the departure airport station to a particular gate at the arrival airport station. To this effect, flight service management can be divided into three separate periods during schedule execution as shown in Fig.~\ref{fig:ADM_FTD}, based upon the management of aircraft operations, namely: Taxi-out, Enroute, and Taxi-in \citep{Midkiff2004}. Taxi-out represents the movement the aircraft from the gate to the runway at the departure airport prior to takeoff, enroute represents the duration and process where the aircraft is airborne as it makes its way to the destination airport, and taxi-in represents transporting the aircraft from the runway to the gate at the arrival airport. As such, the main objective of flight service management is to accurately estimate the duration (i.e block time) between aircraft pushback from the departure gate at the origin airport and parking at the arrival gate at the destination airport. In complement, a secondary objective of flight service management is to minimize, during irregular operations, discretionary holdup (i.e. strategic delay) from aircraft pushback to parking at the departure and arrival airport gates, respectively. Thus, the estimations of turnaround duration and block time present a predictive modeling problem for airline resource capacity (i.e. flight capacity and flight service) management during schedule execution and ADM. 
 
 \subsection{Problem Formulation as an Artificial Neural Network}
 We formulate our PTFM framework for ADM based on the underlying principles of an artificial neural network (ANN). The artificial neural network is a data structure inspired by a framework of biological neurons in living organisms, wherein each neuron is a unit that performs a simple task characterized by responding to an input signal. However, a connected network of neurons (such as the human brain) is capable of completing complex tasks and processes with impeccable speed and accuracy \citep{Rosa2020}. Similar to the biological neural network, an ANN for airline resource capacity management during irregular operations represents a connection of nodes that are analogous to neurons that implicitly describe the physical processes of ADM. To this effect, the ANN is defined by three pertinent characteristics namely: node character, network topology, and learning rules \citep{Zou2008}. 

\begin{figure}[h!]
	\centering
	\includegraphics[width=0.75\textwidth]{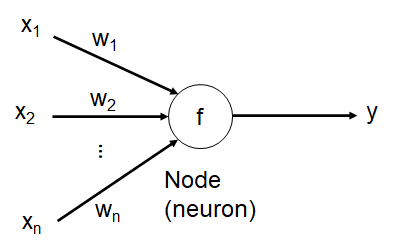}
	\caption{Fundamental model of a single neuron in an artificial neural network}
	\label{fig:neuron}
\end{figure}

\begin{itemize}
    \item \textbf{Node character} represents the information (signal) processing qualities of a node, which are properties such as number of inputs and outputs associated with a node, the appropriate weight for each input and output, and the node's activation function. Fig.~\ref{fig:neuron} shows a generic model for a single neuron or node in an ANN for estimating relevant flight schedule features for optimal airline schedule execution during ADM. The node, which represents a specific flight schedule or disruption feature, obtains multiple inputs from other nodes or flight schedule features that define associated weights describing the strength (or importance) of the flight schedule features with respect to the node. To enable the transmission of information amongst flight schedule features, a transfer function is used to determine the activation of the flight schedule feature (node) when the weighted sum of the inputs from other flight schedule features exceeds a certain threshold value, as shown by the expression in Eqn.~\ref{eqn:ann_tf}
    
    \begin{equation}\label{eqn:ann_tf} 
        \begin{aligned}
            y = f(\sum_{i=1}^{n}{w_{i}x_{i} - T}),
        \end{aligned}
    \end{equation}

    where $y$ is the response of the flight schedule feature (i.e information recipient node), $f$ represents the transfer function, $w_{i}$ is the weight or importance of an input flight schedule feature $x_{i}$, and $T$ is an arbitrary threshold value. The expression for the transfer function in Eqn.~\ref{eqn:ann_tf} can be defined by linear and non-linear functions. However, non-linear functions such as the sigmoid function $S(x)$ \citep{Weisstein2006}, generally expressed as:
    
    \begin{equation}\label{eqn:sig_tf} 
        \begin{aligned}
            S(x) = \frac{1}{1+e^{-x}}
        \end{aligned}
    \end{equation}
    
    are often preferred for modeling real world applications because of their continuous differentiable property \citep{Hunt1971}.   
    
    \begin{figure}[b!]
	    \centering
	    \includegraphics[width=0.99\textwidth]{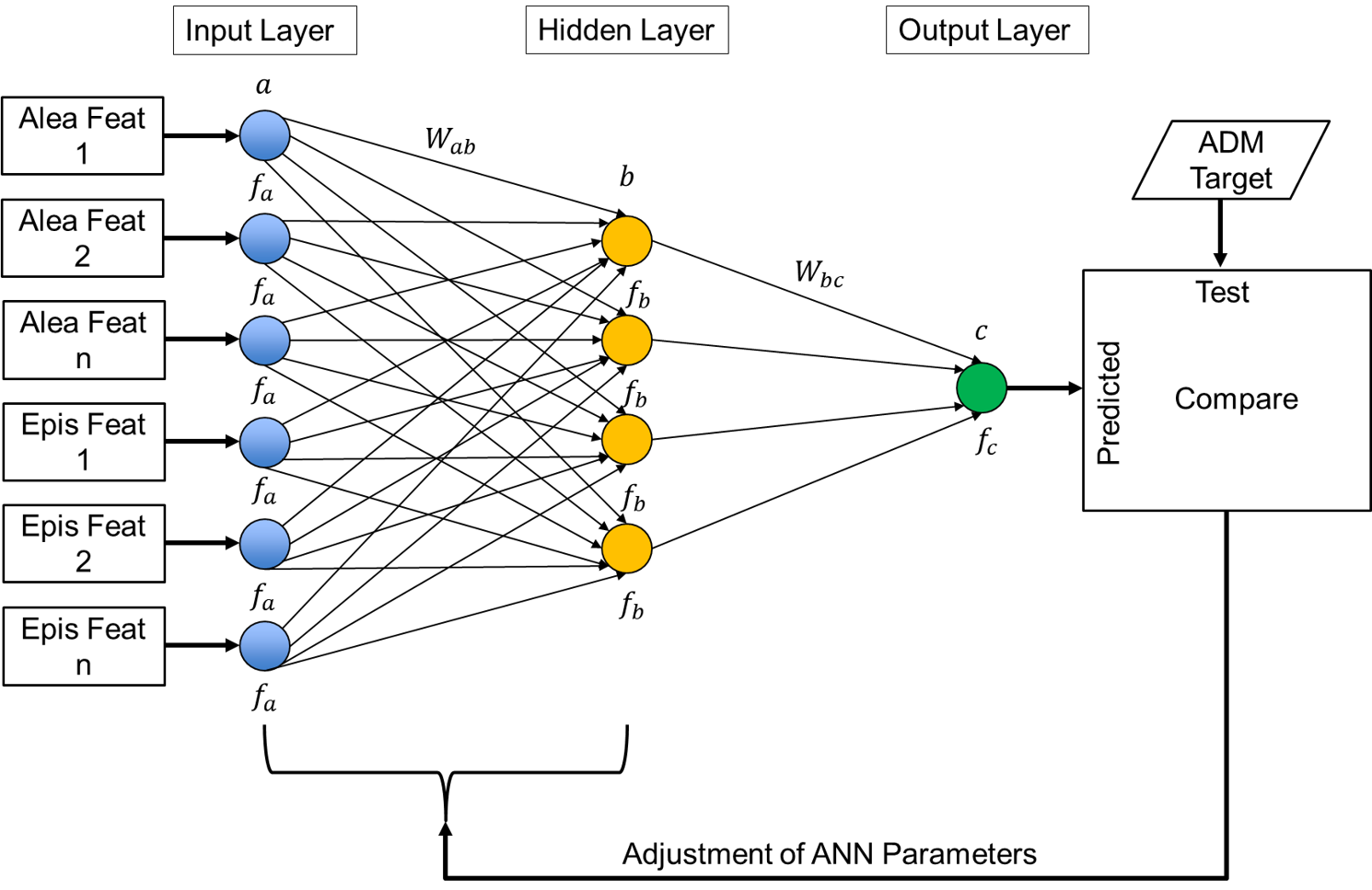}
	    \caption{Artificial neural network framework for estimating target flight schedule features}
	    \label{fig:ANN_topology}
    \end{figure}
    
    \item \textbf{Network topology} of an ANN defines the manner in which nodes representing flight schedule features are organized and connected in a neural network, thereby defining the general architecture of the ANN. For instance, a single hidden layer perceptron  provides an excellent medium for solving linearly separable problems and can also be used to define a Gaussian process \citep{Widrow1990, Neal1996, Lee2018a}. As such, we define the topology of our artificial neural network as a single hidden layer perceptron, based upon the following properties gleaned from outcomes of previous work on exploratory data analysis for ADM \citep{Ogunsina2021}: 
    \begin{enumerate}
        \item Flight schedule and disruption features in the data set are linearly separable; as demonstrated by the existence of orthogonal linear combinations from principal component analysis.
        
        \item The functional process at any stage of airline resource capacity management (such as turnaround) can be defined by a Gaussian process; as demonstrated by the existence of optimal kernel functions and hyperparameters from Gaussian process regression. 

    \end{enumerate}
    
    \begin{figure}[t!]
	    \centering
	    \includegraphics[width=0.75\textwidth]{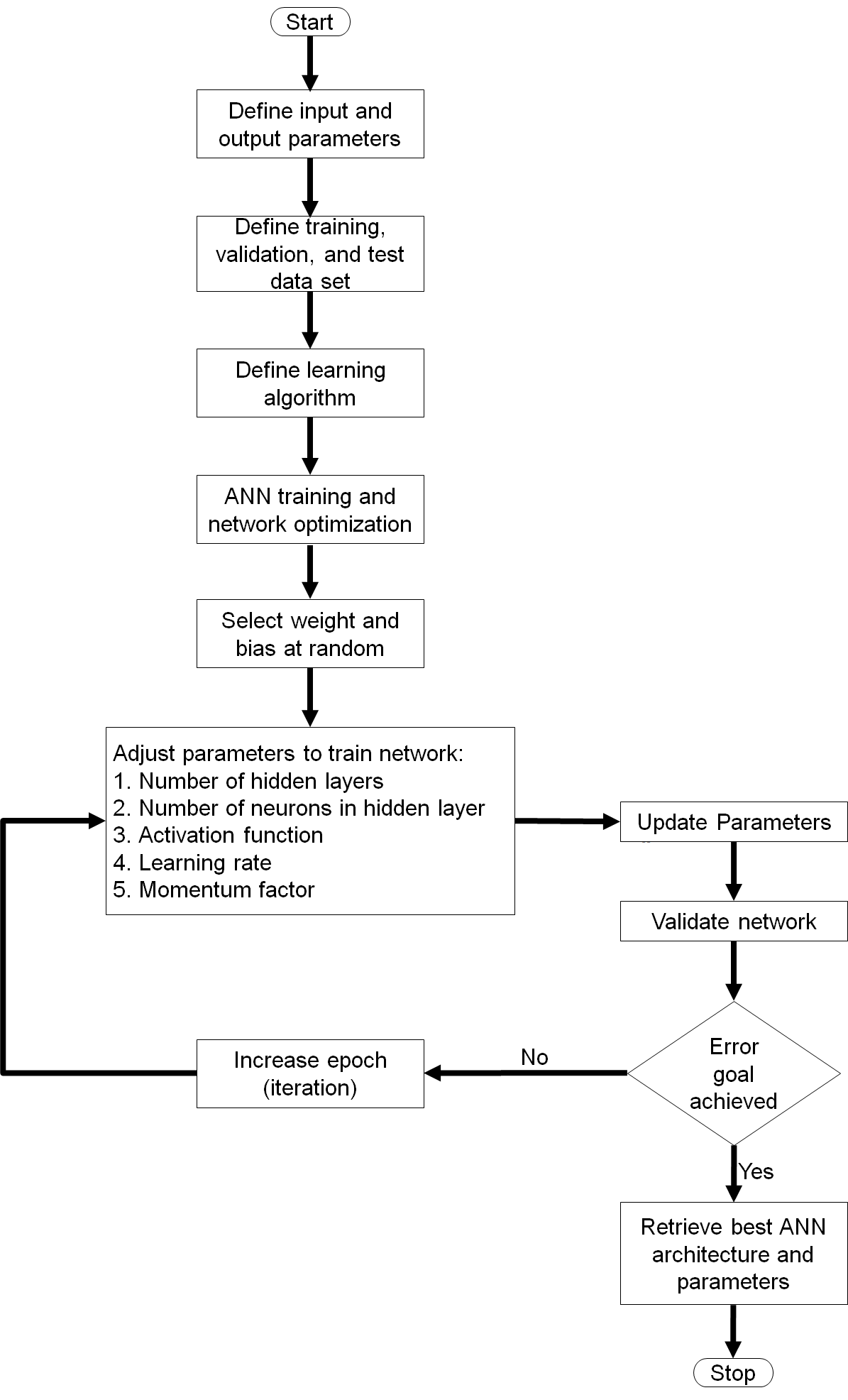}
	    \caption{Flowchart of general process for ANN learning}
	    \label{fig:ANN_learning rules}
    \end{figure}

    Fig.~\ref{fig:ANN_topology} shows the ANN framework for estimating appropriate target features such as turnaround duration for ADM. The ANN framework, shown in Fig.~\ref{fig:ANN_topology}, represents a feedforward perceptron that consists of an input layer $a$ with a set of aleatoric and epistemic flight schedule features \citep{Ogunsina2021} or nodes $f_{a}$, a hidden layer $b$ that contains nodes $f_{b}$ which transmute information from nodes $f_{a}$ in the input layer, and an output layer $c$ with a flight schedule feature or node $f_{c}$ that represents performance measures at any stage of flight schedule execution. For validation of the ANN topology, predicted values of performance measures are compared with actual (realized) values to inform the adjustment of the ANN parameters for improving its predictions.  

    \item \textbf{Learning rules} define the routines that calibrate an ANN by optimizing the parameters of the network topology (or structure) through the use of data that represent instances of flight schedule executions. For calibration, we adopt a supervised learning approach wherein the ANN is trained first before it is applied only when the optimized network topology produces the desired performance from a target feature based upon a set of input features. As such, supervised learning ensures that all possible search rules for obtaining reasonable estimates of target features are inherently cached and readily accessible during irregular operations, thereby reducing the time required for ADM. Fig.~\ref{fig:ANN_learning rules} reveals a generic process for learning a suitable ANN to facilitate the prediction of pertinent features for airline resource management during irregular operations. The process begins by defining input and output flight schedule features appropriate for each stage of airline resource management, and then dividing available instances of flight schedule executions into training and testing categories. Next, a learning algorithm is defined and the optimal parameters for the network topology of the ANN are estimated based upon error correction methods that employ a backpropagation mechanism \citep{Andrew2001}. An error function can be defined as the difference between the flight schedule feature (node) value in the output layer and a corresponding target flight schedule feature value. Let $y_{k,n}$ be the value of the data feature (i.e. node) in the $k^{th}$ output layer at epoch $n$ during training, and $y_{k}^{*}$ be the target value for the data feature in the $k^{th}$ output layer. Thus, the error function is defined as:
    \begin{equation}\label{eqn:ann_erf} 
        \begin{aligned}
            e_{k} = y_{k,n} - y_{k}^{*}
        \end{aligned}
    \end{equation}
    To define an error goal, let $\theta$ be a constant positive value (i.e. learning parameter) that regulates the rate at which the weights of the network are adjusted based upon the following expression:
    \begin{equation}\label{eqn:ann_weightupd} 
        \begin{aligned}
            w_{kj,n+1} = w_{kj,n} - \theta e_{k}x_{j}
        \end{aligned}
    \end{equation}
    Eqn.~\ref{eqn:ann_weightupd} describes the new weight vector at the next epoch, $w_{kj,n+1}$, for an input data feature $x_{j}$ that reduces the error value at epoch $n+1$ based upon error function $e_{k}$. As such, $\theta$ defines the rate at which the ANN learning process shown in Fig.~\ref{fig:ANN_learning rules} converges. 
\end{itemize}

\subsection{Solution Approach for PTFM}
\begin{figure}[h!]
    \centering
	 \includegraphics[width=0.6\textwidth]{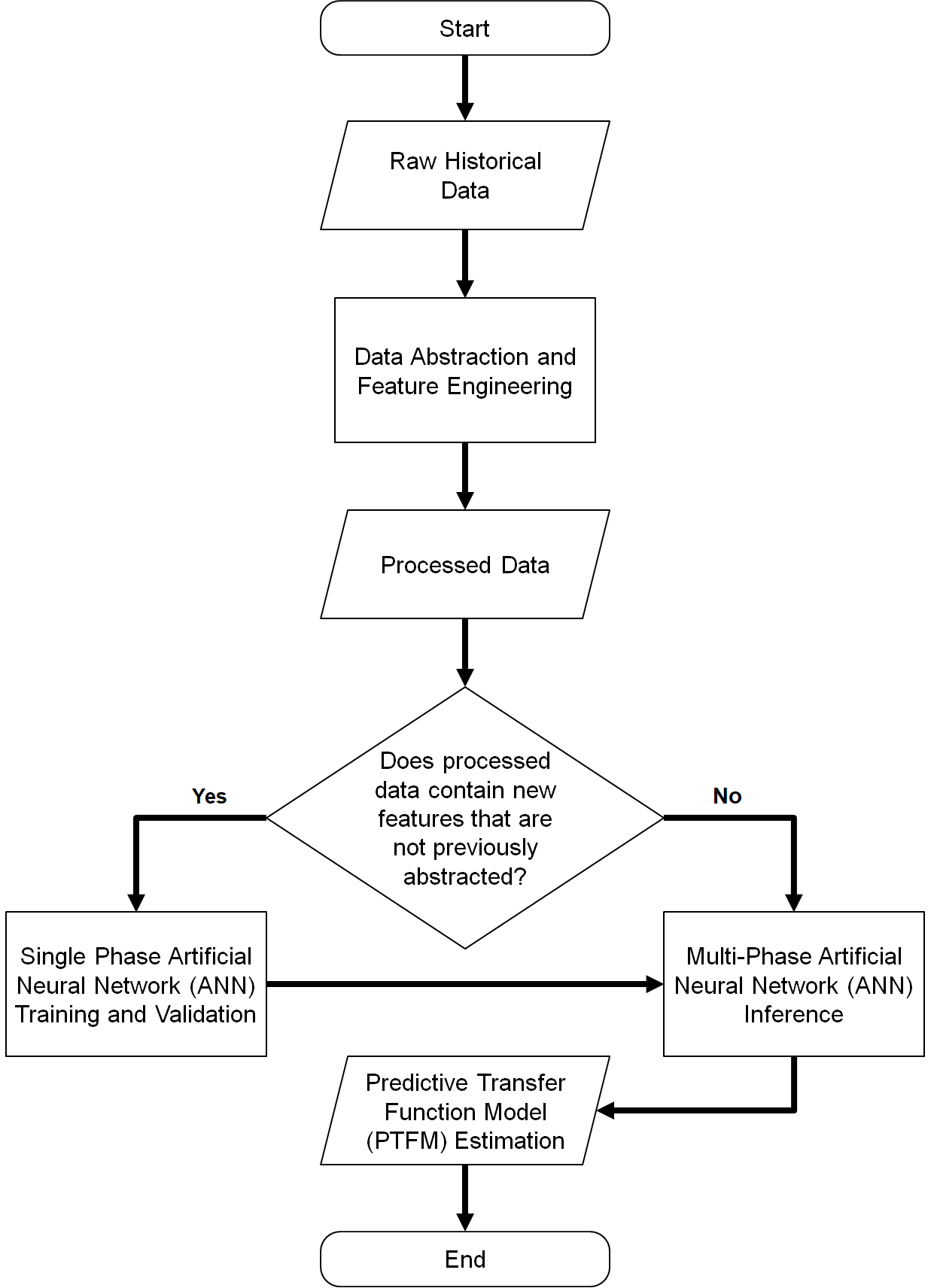}
	 \caption{Component assembly approach for automatic estimation of performance measures for disruption management}
	 \label{fig:PTFM_Solution}
\end{figure}
We employ a component assembly process \citep{Cao2005} to create artificial neural networks that predict different target features for flight capacity management and flight service management at all phases of ADM. Fig.~\ref{fig:PTFM_Solution} shows our solution approach for automatic PTFM estimation of performance measures for ADM. The approach begins by preprocessing raw data into readily decipherable data formats that are suitable for ANN learning algorithms, by employing the abstraction and techniques described in \cite{Ogunsina2021}. Next, we learn the optimal network topology of each ANN for predicting target flight schedule features at each stage of airline resource capacity management, before performance measures at each stage are estimated through a combination of flight schedule features inferred from the optimal ANNs.   

For the remainder of this section, we discuss the methods employed to define and learn optimal structures of separate ANNs that represent components of the PTFM for ADM. Next, we describe a simple set of overarching rules for retrieving PTFM performance estimates for flight capacity management and flight service management during irregular airline operations.   

\subsection{Single Phase ANN Development} \label{single_phase_dev}
Single phase ANN development represents the creation of artificial neural networks for each of the three individual phases of ADM namely: proactive ADM before schedule execution (i.e. tactical ADM), reactive ADM during schedule execution (i.e. operational ADM), and proactive ADM after schedule execution (i.e. strategic ADM). By applying guidelines defined in Fig.~\ref{fig:ANN_learning rules}, we discuss key routines and rationale for developing customizable ANNs for each phase of ADM.  

\subsubsection{Defining Inputs and Targets}
For each of the three phases of ADM, there exist relevant flight schedule features in a historical airline scheduling and operations data set that are of significant interest to AOCC practitioners. With respect to the processes involved before flight schedule execution, tactical ADM aims to eradicate any inconvenience to passengers that may arise during flight capacity management of irregular operations prior to aircraft pushback from the departure airport gate. As such, turnaround duration represents an appropriate target flight schedule feature for a feedforward ANN for tactical ADM. Operational ADM seeks to ensure that the departure and arrival flight chronology retrospectively set by airlines (and sanctioned by other air transportation stakeholders) remains valid during schedule execution. To that effect, binary indicators for on-time flight arrival and arrival within $14 mins$ of original schedule (i.e. $A0$ and $A14$ respectively) represent suitable target flight schedule features for a feedforward ANN for operational ADM. Complementary to tactical ADM, strategic ADM aims to eradicate operational delays and inconsistencies during flight service management of irregular operations after aircraft pushback from the departure airport gate. Thus, block time represents a suitable target flight schedule feature for an ANN for strategic disruption management. From a feature abstraction perspective, the target flight schedule features for all three phases of ADM represent epistemic flight schedule features that are activated due to lack of knowledge of the exact impact of their applied alternatives (by AOCC practitioners) during ADM. 

Inputs for the artificial neural networks for each of the three phases of ADM are defined based upon appropriate aleatoric and epistemic features that are observable at different stages of flight schedule execution. As such, all determinate aleatoric features (i.e. features representing flight date, origin airport, destination airport, number of passengers, etc.) are viable ANN inputs that indicate the uniqueness of a particular flight schedule with respect to the airline route network. Furthermore, indeterminate aleatoric features (such as features representing IATA delay codes) are necessary ANN inputs for estimating on-time performance for reactive ADM during schedule execution, and define the uniqueness of a particular flight schedule with respect to the disruptions encountered in the airline route network. Lastly, epistemic features provide inputs that indicate the uniqueness of the disruption resolution (or \textit{rule-of-thumb}) applied by a human specialist in the AOCC (i.e. intelligent agent) during irregular operations and disruption management.  

\subsubsection{Data Segmentation for Learning}
Prior to learning the optimal parameters of the network topology of ANNs for each phase of disruption management, it is necessary to define the data set that will inform learning and validation based on the general objectives for each phase. There are two distinct chunks (or subsets) of data that make up the full data set provided by a major U.S. airline, namely: \textit{non-disrupted} and \textit{disrupted} data sets. The \textit{non-disrupted} data set represents all instances of flight schedules in the U.S. airline's network that executed without any disruptions between September 2016 and September 2017, and the \textit{disrupted} data set represents all instances of flight schedules in the airline network that executed with disruptions (i.e. delays) during the same time frame. As such, two separate ANNs that determine the turnaround duration for tactical ADM are created by using the \textit{non-disrupted} data set and \textit{disrupted} data set respectively. Similarly, two separate feedforward ANNs that define the estimation of actual block time duration for strategic ADM are developed from the \textit{non-disrupted} and \textit{disrupted} data sets respectively. For evaluation of on-time performance for operational ADM, two separate ANNs that predict on-time aircraft arrival and aircraft arrival within $14 mins$ at the destination airport gate, respectively, are created using the \textit{disrupted} data set. Thus, a total of six distinct feedforward ANNs are created across all three phases of ADM. The appropriate data set for developing each of the ANNs is further partitioned, via a random seed of 42, such that 70\% of the instances of flight schedule execution in the data set are randomly selected to train the ANN, and the remaining 30\% of flight schedule executions are used to test the validity the trained ANN. 

\subsubsection{ANN Learning and Validation}
Upon defining the inputs, targets and appropriate data sets for learning the ANNs for each phase of ADM, the next step in the single phase ANN development process is to find optimal parameters for the network topology that defines the ANNs. Thus, we define the network topology for each of the six ANNs as the single hidden layer perceptron shown in Fig.~\ref{fig:ANN_topology}. All nodes in input and output layers of the ANNs are activated by linear functions, while nodes in the hidden layer are activated by nonlinear functions described later in this section. The optimal size (i.e. number of nodes) of the hidden layer is typically between the size of the input and size of the output layers \citep{Heaton2015}. Hence, we set the number of nodes in the hidden layer of the perceptron as the ceiling value of the average of the number of nodes in the input layer and the number of nodes in the output layer, for each of the six ANNs. 

\begin{enumerate}
\item \textit{Learning}:
ANNs for tactical and strategic ADM have continuous targets (i.e. turnaround duration and block time respectively). As such, learning optimal weights for these ANNs presents a regression problem. Thus, we define the error function for adjusting the weights of the single hidden layer perceptron as the Huber loss function ($H$) \citep{Huber1964}, expressed as follows:

\begin{equation}\label{eqn:huber_loss1} 
    \begin{aligned}
        H(x,y) = \frac{1}{n}\sum_{i}^{n}{z_{i}}
    \end{aligned}
\end{equation}
such that $z_{i}$ is given by:
\begin{align} \label{eqn:huber_loss2}
    z_{i} = 
    \begin{cases}
    0.5(x_{i}^{2}+y_{i}^{2}) &\quad\text{if} \quad |x_{i} - y_{i}|< 1, \\
    |x_{i}-y_{i}|-0.5 &\quad\text{otherwise} \\
    \end{cases}
\end{align}
where $x_{i}$ and $y_{i}$ represent each $i^{th}$ entry in the input feature space and target feature space of the training data set respectively, and $n$ represents the total number of instances of flight schedule executions in the training data set. As shown in Eqns.~\ref{eqn:huber_loss1} and \ref{eqn:huber_loss2}, the Huber loss function creates an error goal that uses a squared term if the absolute element-wise error is below 1 and an $L1$ term otherwise. This makes it amenable (i.e. less sensitive) to accommodating for outliers during ANN training. We employ the logarithm of the sigmoid function to activate the nodes in hidden layers of the ANNs for tactical and strategic ADM during the search for optimal network parameters.

ANNs for operational ADM have binary targets (i.e. 0 or 1) for whether or not an aircraft arrives at the destination airport gate at a particular time. As such, learning optimal weights for these ANNs presents a classification problem. To that effect, we define the error function for adjusting the weights of the single hidden layer network topology for operational ADM as a function that measures the binary cross entropy ($BCE$) \citep{Mannor2005} between the target and output layer value, expressed as follows:

\begin{equation}\label{eqn:bce1} 
    \begin{aligned}
        BCE(x,y) = \frac{1}{n}\sum_{k}^{n}{b_{k}}
    \end{aligned}
\end{equation}
such that $b_{k}$ is given by:
\begin{equation}\label{eqn:bce2} 
    \begin{aligned}
        b_{k} = -[y_{k}\log{x_{k}}+(1-y_{k})\log{(1-x_{k})}]
    \end{aligned}
\end{equation}
where $x_{k}$ is the value (i.e. probability of predicting the $k^{th}$ target value) from the node in the output layer of the ANN, $y_{k}$ is the $k^{th}$ target value adjusted by the sigmoid function and $n$ represents the total number of instances of flight schedule executions in the training data set. The nodes in the hidden layer of ANNs for operational ADM are activated through the softplus function $s(x)$, expressed in Eqn.~\ref{eqn:softplus}, to obtain the weights that optimally define the network topology during training.
\begin{equation}\label{eqn:softplus} 
    \begin{aligned}
        s(x) = \log{(1+e^{x})}
    \end{aligned}
\end{equation}
We apply the popular adaptive moment estimation (Adam) learning algorithm \citep{Kingma2015} to find the optimal parameters of each of the six ANNs, by ensuring that the error goal for training the respective ANNs remains constant after a considerable amount of epochs.

\item \textit{Validation}: After learning the optimal weight parameters of the network topology for each of the six ANNs for separate phases of ADM, we verify the credibility of the estimations from the neural networks. The ANNs that address the regression problems for tactical and strategic ADM are validated by comparing the actual turnaround and block time duration values from appropriate test (unseen) data sets with the turnaround and block time duration values predicted by the ANNs, based upon corresponding input flight schedule features from the test data sets. As such, we employ the root mean square error (RMSE) to indicate the absolute fit of a generic ANN's predicted values with observed test data values. This provides a measure of the standard deviation of the unexplained (residual) variance in the predictive capacity of the ANN \citep{Shekhar2008}. RMSE also has the useful property of being in the same units as the target values from the test data and is expressed as follows:

\begin{equation}\label{eqn:rmse} 
    \begin{aligned}
        RMSE = \sqrt{\sum_{i=1}^{n}{\frac{(\hat{y}_{i}^{2}-y_{i}^{2})}{n}}}
    \end{aligned}
\end{equation}

where $y_{i}$ represents a target value in a test data set, $\hat{y_{i}}$ indicates the corresponding predicted value from the ANN, and $n$ represents the total number of samples in the test data set.

For ANNs that solve the classification problem for operational ADM, we use a special parameter called area under the receiver operating characteristic (ROC) curve (or AUC for short) \citep{Bradley1997, Myerson2001} to test the validity of the ANNs. The ROC curve is a two-dimensional graph that reveals the performance of a classification model for any classification threshold. This curve plots two parameters namely: true positive rate and false positive rate. True positive rate, also known as recall or sensitivity, is a function of correct ANN predictions of on-time arrival (i.e. $A0$ or $A14$) from the test data set that are identified (i.e true positive or TP) and incorrect ANN predictions of $A0$ or $A14$ from the test data set that are rejected (i.e. false negative or FN). False positive rate, also known as fallout, is a function of incorrect ANN predictions of on-time arrival that are identified from the test data set (i.e. false positive or FP) and correct ANN predictions of on-time arrival from the test data set that are rejected (i.e. true negative or TN). As such, recall and fallout are expressed as follows:
\begin{equation}\label{eqn:recall} 
    \begin{aligned}
        recall = \frac{TP}{TP+FN}
    \end{aligned}
\end{equation}

\begin{equation}\label{eqn:recall} 
    \begin{aligned}
        fallout = \frac{FP}{FP+TN}
    \end{aligned}
\end{equation}

Thus, AUC measures the entire two-dimensional area underneath an ROC curve from (0,0) to (1,1). AUC values range from 0 to 1, such that an ANN whose predictions are 100\% wrong has an AUC of 0.0 while an ANN whose predictions are 100\% correct has an AUC of 1.0.
\end{enumerate}

\subsection{Multi-Phase ANN Inference}

\begin{sidewaysfigure}[htbp!]
    \centering
	 \includegraphics[width=0.99\textwidth]{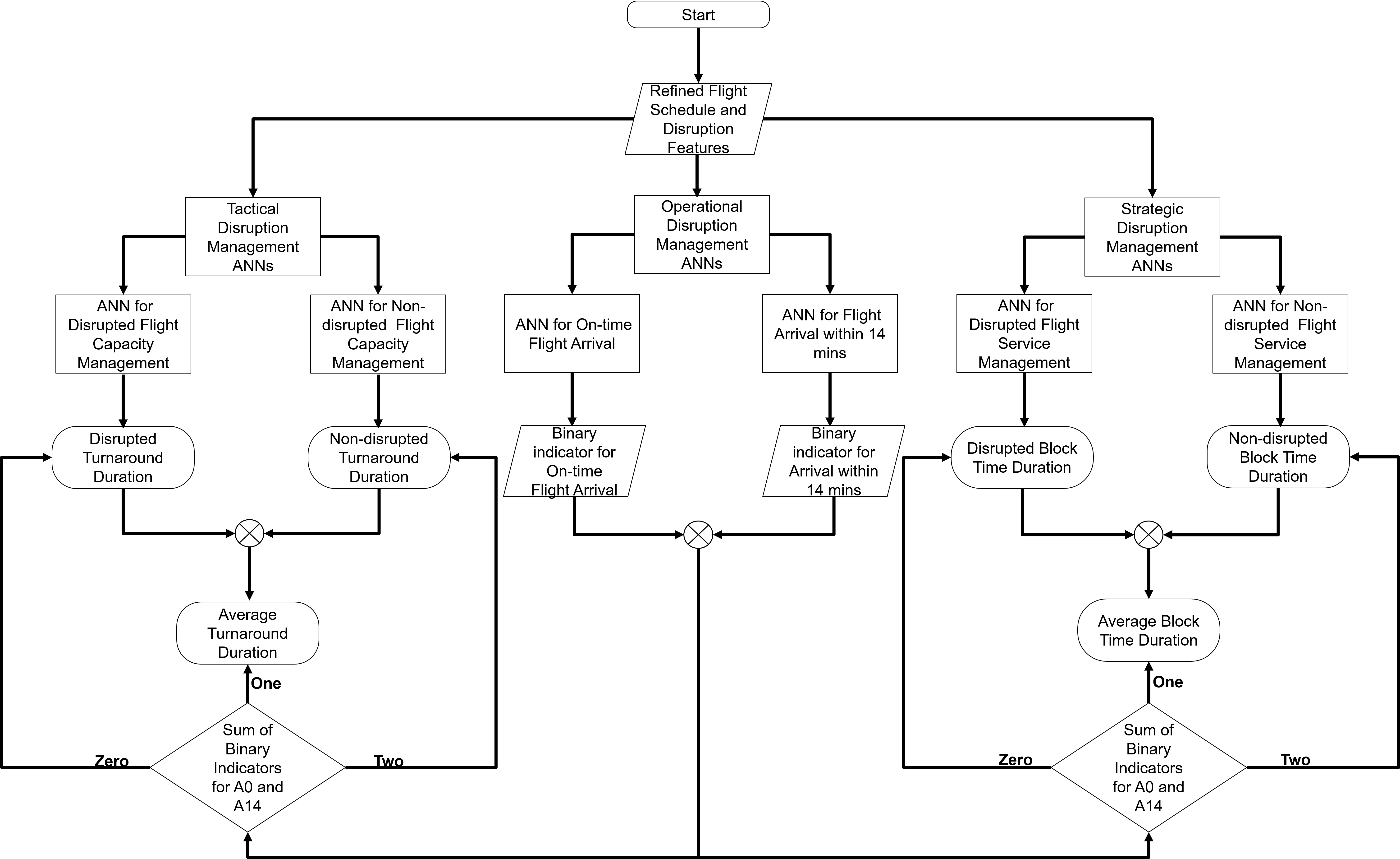}
	 \caption{Flowchart of parallel ensemble approach for PTFM estimation}
	 \label{fig:PTFM_Inference}
\end{sidewaysfigure}

Multi-phase ANN inference represents the appropriate estimation of turnaround duration and block time for both stages of airline resource management during irregular operations and disruption management. As mentioned in Section \ref{single_phase_dev}, the estimation of separate features of interest at each phase of ADM relies on multiple ANNs. As such, we apply a parallel ensemble approach \citep{Zhou2012} to obtain quality estimations of turnaround duration and block time. Thus, the component ANNs for each phase of ADM are independently learned in parallel and their corresponding predictions are combined via a custom bootstrap aggregating (i.e. bagging) procedure. The bootstrap aggregating approach leverages the independence between the component ANNs (i.e. base learners) to ensure that the overall error is significantly reduced \citep{Zhou2012}, when combining multiple independent ANN predictions to retrieve the PTFM estimates of turnaround duration and block time. 

Fig.~\ref{fig:PTFM_Inference} reveals the parallel ensemble approach for multi-phase ANN inference to obtain PTFM estimates of turnaround duration and block time for airline resource management and flight schedule execution during irregular operations. The bagging process begins by obtaining necessary refined data features for a disrupted flight schedule, which represent inputs of separate ANNs for tactical, operational, and strategic ADM. Next, the predictions of binary indicators for on-time flight arrival and flight arrival within $14 mins$ of scheduled arrival, at the destination airport gate, are retrieved through inference of the respective ANNs for operational ADM. The inferred values of the binary indicators for both measures of on-time performance (i.e. $A0$ and $A14$) are subsequently summed, and used to retroactively estimate turnaround duration and block time through the inference of respective ANNs for tactical and strategic ADM. A sum of zero (for $A0$ and $A14$) implies that a disrupted flight does not arrive as originally scheduled nor does it arrive within $14 mins$ of the scheduled time. A sum of one for $A0$ and $A14$ implies that the aircraft on a disrupted flight either arrives precisely as originally scheduled or within $14 mins$ beyond the original scheduled arrival time at the destination airport gate. A sum of two for $A0$ and $A14$ indicates that a disrupted flight either arrives exactly as originally scheduled or within $14 mins$ earlier than the precise arrival time scheduled for parking at the destination airport gate.   

If the sum of the values for $A0$ and $A14$ is zero, then the turnaround duration and block time predictions from the tactical and strategic ADM ANNs, trained by using the \textit{disrupted} data set, are set as the PTFM estimation of turnaround duration and block time respectively. If the sum of the values for $A0$ and $A14$ is one, then the averages of the turnaround duration and block time predictions from tactical and strategic ADM ANNs that are trained by using the \textit{disrupted} and \textit{non-disrupted} data sets respectively, are set as the PTFM estimates of turnaround duration and block time. If the sum of the predicted values for $A0$ and $A14$ is two, then the turnaround duration and block time predictions from the tactical and strategic ADM ANNs, trained by using the \textit{non-disrupted} data set, are set as the PTFM estimation of turnaround duration and block time respectively. Thus, we estimate the tactical delay incurred during flight capacity management as the difference between the turnaround duration predictions retrieved from the ANNs for tactical ADM, trained by using the \textit{disrupted} and \textit{non-disrupted} data sets respectively. Congruently, the strategic delay period accrued during flight service management is estimated as the difference between the block time predictions retrieved from respective ANNs for strategic ADM, calibrated through the \textit{disrupted} and \textit{non-disrupted} data sets.

\section{Computational Setup} \label{setup}

\begin{table}[b!]
\begin{center}
\caption{Epistemic features for multi-phase ANN inference in PTFM} \label{tab:PTFM_Inputs}
\begin{tabular}{*3c}
  \hline
	\textbf{ADM Phase}   & \textbf{Epistemic Input Features} & \textbf{Target Feature}\\ \hline
	 Tactical  & \textit{ADJST\_TURN\_MINS} & \textit{ACTL\_TURN\_MINS} \\ 
	 Operational   & \textit{shiftper\_sched\_PB, shiftper\_sched\_GP, DOT\_DELAY\_MINS}  & \textit{A0, A14} \\
	 Strategic  & \textit{shiftper\_actl\_PB, shiftper\_actl\_GP, actl\_enroute\_mins}  & \textit{actl\_block\_mins}\\ 
    \hline
\end{tabular}
\end{center}
\end{table}

We now discuss the computational setup for the predictive transfer function model for a characteristic computational and intelligent agent (i.e functional role) in the AOCC. Table~\ref{tab:PTFM_Inputs} summarizes specific epistemic input flight schedule features and corresponding target features in all ANNs for separate phases of ADM. These features represent inputs and targets that are applicable for separate classes of ANNs for each phase of ADM, and are adopted based upon the results observed from exploratory data analysis \citep{Ogunsina2021}. Furthermore, all determinate aleatoric flight schedule features represent additional generic inputs for all ANNs for each phase of ADM. In complement, indeterminate aleatoric features (i.e. disruption features) represent additional specific inputs for ANNs that define the PTFMs for different functional roles in the AOCC during operational ADM. The definitions of aleatoric and epistemic features used for PTFM demonstration in this paper can be found in Table~\ref{tab:Definitions}. Following appropriate data preprocessing and segmentation, all ANNs are subsequently implemented through parallel learning (for 15,000 epochs) and then inference in the Python programming language. The learning and inference processes are significantly accelerated through computations via the \textit{PyTorch} software \citep{Ketkar2017} running on an NVIDIA GTX 1080Ti graphics card \citep{PyTorchCommunity2016}. 

\begin{longtable}{*3c}
\caption{Definitions of aleatoric and epistemic features used for PTFM demonstration} \label{tab:Definitions} \\ \hline
	\textbf{Data Feature} & \textbf{Description} & \textbf{Feature Class}\\
	\hline
	\textit{A0} &  Binary indicator for on-time arrival & Epistemic\\
	\textit{A14} &  Binary indicator for on-time arrival within $14 mins$ & Epistemic\\
	\textit{actl\_block\_mins} &  Actual block time duration & Epistemic\\ 
	\textit{actl\_enroute\_mins} &  Actual flight period in the air & Epistemic\\ 
	\textit{ACTL\_TURN\_MINS} &  Actual turnaround period & Epistemic\\
	\textit{ADJST\_TURN\_MINS} &  Adjusted turnaround period & Epistemic\\ 
	\textit{DOT\_DELAY\_MINS} &  Total arrival delay & Epistemic\\
	\textit{doy} & Day of the year & Determinate aleatoric\\ 
	\textit{dest\_x\_dir} & Destination airport location in spherical X coordinate & Determinate aleatoric\\ 
	\textit{dest\_y\_dir} & Destination airport location in spherical Y coordinate & Determinate aleatoric\\ 
	\textit{dest\_z\_dir} & Destination airport location in spherical Z coordinate & Determinate aleatoric\\ 
	\textit{HD06} &  ATC Hold for bad weather at departure station & Indeterminate aleatoric\\ 
	\textit{ONBD\_CT} & Total number of passengers onboard flight  & Determinate aleatoric\\ 
	\textit{orig\_x\_dir} & Origin airport location in spherical X coordinate & Determinate aleatoric\\ 
	\textit{orig\_y\_dir} & Origin airport location in spherical Y coordinate & Determinate aleatoric\\ 
	\textit{orig\_z\_dir} & Origin airport location in spherical Z coordinate & Determinate aleatoric\\ 
	\textit{route\_dist} & Spherical distance between origin and destination airports & Determinate aleatoric\\ 
	\textit{route\_originator\_flag} &  Binary indicator for first flight of the day & Determinate aleatoric\\ 
	\textit{shiftper\_actl\_PB} &  \% work shift completed at actual pushback time & Epistemic\\ 
	\textit{shiftper\_actl\_GP} &  \% work shift completed at actual gate parking time & Epistemic\\
	\textit{shiftper\_sched\_PB} &  \% work shift completed at scheduled pushback time & Epistemic\\ 
	\textit{shiftper\_sched\_GP} &  \% work shift completed at scheduled gate parking time & Epistemic\\ \hline
\end{longtable}

\section{PTFM Analysis and Results} \label{results}
This section is comprised of two separate parts, which discuss different analyses and results from the development and implementation of the PTFM respectively. The first part on PTFM development (i.e. single phase outcome) demonstrates the evaluation of the optimal ANN topology for the PTFM of the Weather functional role in the AOCC at each phase of ADM. The second part on PTFM implementation (i.e. multi-phase outcome) demonstrates the assessment of PTFM estimations obtained through the combination of predictions from multiple optimal ANNs for ADM. For both parts, we demonstrate the analysis of the PTFM by using weather-delayed flight schedules as the \textit{disrupted} data set to represent the Weather functional role in the AOCC. 

\begin{figure}[b!]
    \centering
	  \includegraphics[width=0.99\textwidth]{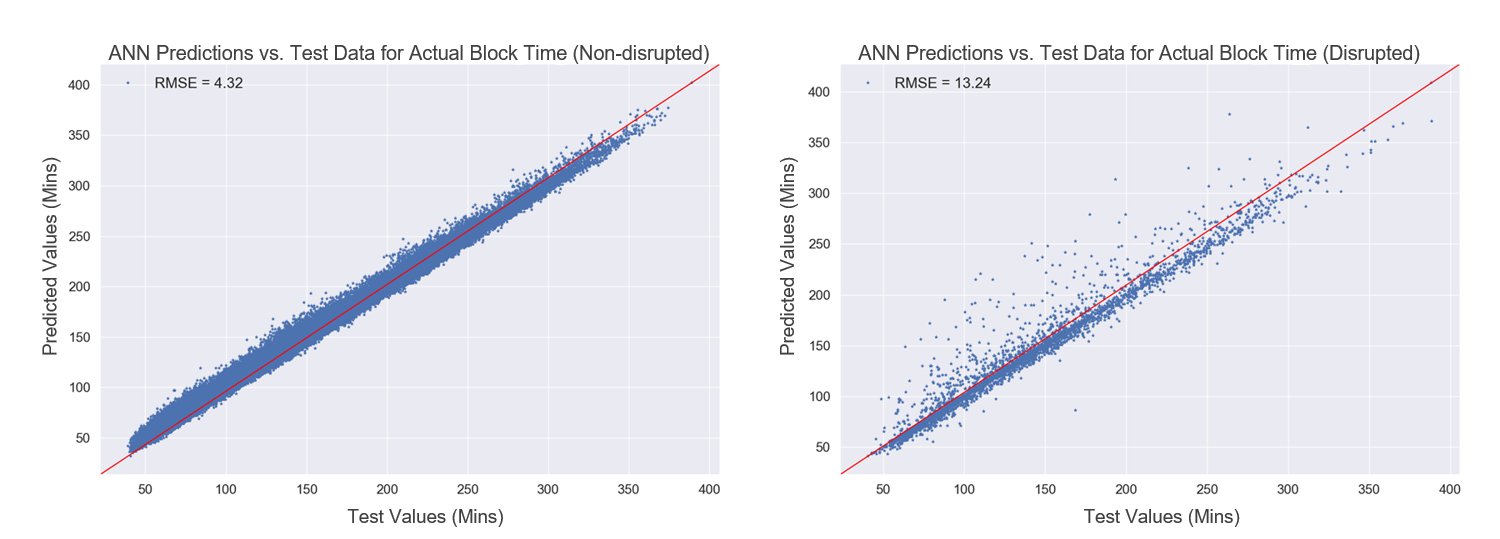}
	    \caption{ANN predictions vs. test data for strategic ADM}
	  \label{fig:BT_results}
\end{figure}

\subsection{Single Phase Results}

Fig.~\ref{fig:BT_results} shows the plots of the predictions of actual block time duration versus the observed block time duration for instances of non-disrupted flight schedules and delayed flight schedules due to weather disruptions respectively. The plot on the left in Fig.~\ref{fig:BT_results} represents the predicted block time from the ANN for strategic ADM, learned by using the \textit{non-disrupted} data set of 430,000 training samples and 186,000 test samples. The left plot in Fig.~\ref{fig:BT_results} reveals that the predictive capacity of the ANN for non-disrupted flight service management during strategic ADM has a standard deviation of unexplained variance (i.e. RMSE) of $4.23 mins$ between the aircraft pushback at the origin to parking at the destination gate. The plot on the right in Fig.~\ref{fig:BT_results} shows the predicted block time from the ANN for strategic ADM. The optimal network topology of the ANN was obtained by using the smaller set of \textit{disrupted} data (8,861 training samples and 3,798 test samples) due to inclement weather events. Unlike the left plot in Fig.~\ref{fig:BT_results}, the plot on the right reveals a lower predictive capacity for disrupted flight service management during strategic ADM of weather-related delays, with an RMSE of about $13 mins$ between aircraft pushback and aircraft gate-parking for a specific flight.   

\begin{figure}[h!]
    \centering
	   \includegraphics[width=0.99\textwidth]{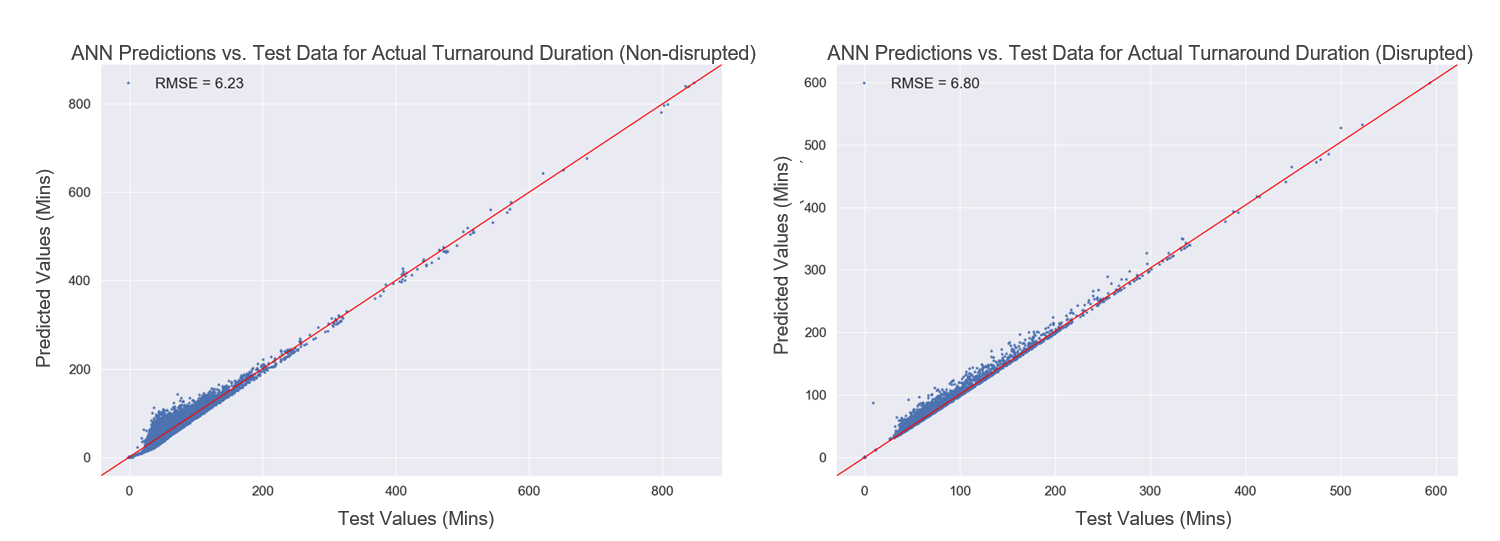}
	   \caption{ANN predictions vs. test data for tactical ADM}
	   \label{fig:TT_results}
\end{figure}
    
Fig.~\ref{fig:TT_results} represents the plots of the predictions of actual turnaround duration versus the observed turnaround duration for instances of non-disrupted flight schedules and delayed flight schedules respectively. Similar to Fig.~\ref{fig:BT_results}, the plot on the left in Fig.~\ref{fig:TT_results} represents the predicted turnaround duration from the ANN for tactical ADM, learned by using the \textit{non-disrupted} data set of instances of flight schedule executions. Hence, the left plot in Fig.~\ref{fig:TT_results} reveals that the predictive capacity of the ANN for non-disrupted flight capacity management during tactical ADM has an RMSE of $6.23 mins$ between passenger boarding of an aircraft to pushback of the aircraft from the origin airport gate. 
The plot on the right in Fig.~\ref{fig:TT_results} shows the predicted turnaround duration from the ANN for tactical ADM, learned by using the \textit{disrupted} data set of instances of delayed flight schedules due to inclement weather events. Unlike Fig.~\ref{fig:BT_results}, the plot on the right in Fig.~\ref{fig:TT_results} reveals a relatively similar predictive capacity for disrupted flight capacity management during tactical ADM of weather-related delays, with an RMSE of $6.80 mins$ between aircraft boarding and aircraft pushback for a specific flight. 
    
\begin{figure}[ht!]
    \centering
	   \includegraphics[width=0.99\textwidth]{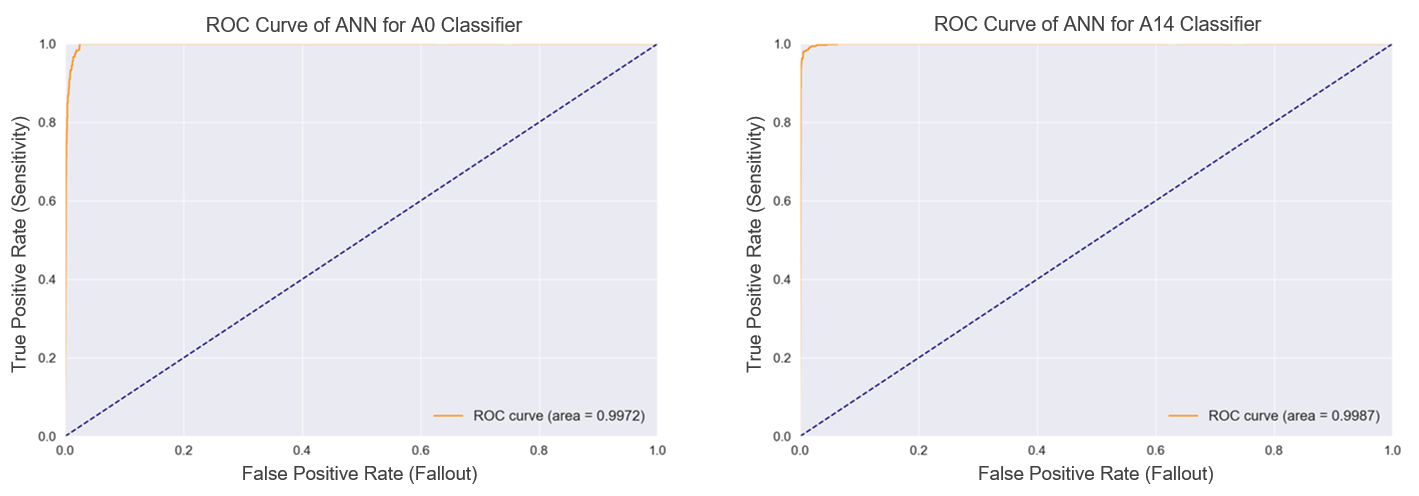}
	   \caption{ROC plots of ANN classifiers for operational ADM}
	   \label{fig:A0_A14_results}
\end{figure}
    
The red diagonal line in Fig.~\ref{fig:BT_results} and Fig.~\ref{fig:TT_results} represents the line of perfect  prediction, such that data points on this line indicate a perfect prediction of the observed test data value by the ANN. As such, Fig.~\ref{fig:TT_results} reveals near-perfect predictions of turnaround duration that are over $400 mins$ (i.e. outliers) for both non-disrupted and disrupted flight schedules observed, respectively, in the test data sets.

\begin{sidewaystable}
\begin{center}
\caption{Model performance summary of ANNs for all functional roles in AOCC} \label{tab:ANN_summaries}
\begin{tabular}{*7c}
  \hline
	\textbf{Functional Role} & \textbf{Training Data Samples} & \textbf{Test Data Samples} &\textbf{Block Time RMSE (mins)} & \textbf{Turnaround RMSE (mins)} & \textbf{A0 AUC} & \textbf{A14 AUC} \\ \hline
	\textbf{\textit{Customer Hold}} & 32,809 & 14,061 & 5.97 & 7.15 & 1.0000 & 1.0000 \\ 
	\textbf{\textit{Dispatch CSC}} & 12,228  & 5,240 & 8.79 & 6.58 & 1.0000  & 0.9996 \\
	\textbf{\textit{Flight Operations}} & 25,459  & 10,911 & 7.04 & 6.93 & 1.0000 & 1.0000\\
	\textbf{\textit{Fuel Management}} & 3,389 & 1,452  & 8.02 & 5.88 & 0.9946 & 0.9929\\
	\textbf{\textit{Ground Operations}} & 117,863 & 50,512 & 6.58  & 5.87 & 1.0000 & 1.0000\\ 
	\textbf{\textit{Inflight}} & 55,611  & 23,833 & 6.99 & 5.63 & 1.0000 & 0.9999 \\
	\textbf{\textit{Maintenance}} & 23,463 & 10,055 & 7.21 & 6.70 & 1.0000 & 1.0000\\
	\textbf{\textit{NAS}} & 15,851 & 6,793 & 8.97 & 6.85 & 0.9998 & 0.9999\\
	\textbf{\textit{Security}} & 2,069 & 886 & 7.34 & 6.59 & 0.9849 & 0.9895 \\
	\textbf{\textit{Technology}} & 6,267 & 2,686  & 7.47 & 6.12 & 1.0000 & 1.0000\\
    \textbf{\textit{Weather}} & 8,861 & 3,798 & 13.24 & 6.80 & 0.9972 & 0.9987\\
    \hline
\end{tabular}
\end{center}
\end{sidewaystable}
    
Fig.~\ref{fig:A0_A14_results} shows the ROC plots of the separate ANNs for operational ADM, which predict if a specific disrupted flight arrives exactly on-time and within $14 mins$ of the scheduled arrival time, respectively. Thus, both ANNs, described by Fig~\ref{fig:A0_A14_results}, are learned by using the \textit{disrupted} data set of instances of delayed flight schedule executions. The blue dashed diagonal line in both plots in Fig.~\ref{fig:A0_A14_results} represents perfect chance. That is, an ANN for operational ADM that makes predictions based upon the blue diagonal line has no better odds of detecting $A0$ and $A14$ than a random coin flip. The solid orange curve from the plot on the left in Fig.~\ref{fig:A0_A14_results} represents the ROC curve of the ANN classifier for assessing on-time arrival (i.e. $A0$) of a disrupted flight schedule. The area under the ROC curve (i.e. AUC) is 0.9972, which indicates that the ANN can correctly classify the on-time arrival status of a disrupted flight schedule over 99\% of the time. Similar to the left plot, the solid orange curve from the plot on the right in Fig.~\ref{fig:A0_A14_results} reveals the ROC curve of the ANN classifier for assessing arrival within $14 mins$ (i.e. $A14$) of a disrupted flight schedule. The area under the orange ROC curve, from the plot on the right in Fig.~\ref{fig:A0_A14_results}, is 0.9987, indicating that the ANN can correctly identify if a disrupted flight schedule arrives within $14 mins$ of the original schedule over 99\% of the time.  
    
Table~\ref{tab:ANN_summaries} shows the summary of the model performance of separate ANNs in estimating appropriate flight schedule features for all functional roles \citep{Ogunsina2021} in the AOCC during disruption management. As shown in Table~\ref{tab:ANN_summaries}, all ANNs for operational ADM have near perfect performance with AUC values of approximately 1. The highest RMSE value (i.e. $7.15 mins$) for predicting turnaround duration for disrupted flight capacity management during schedule execution (i.e. tactical ADM) was attained by the ANN for the Customer Hold functional role in the AOCC. Table~\ref{tab:ANN_summaries} also shows that the highest RMSE value for predicting block time for disrupted flight service management across all functional roles during schedule execution (i.e. strategic ADM) is less than $14 mins$, and attained by the ANN for the Weather functional role in the AOCC. Thus, the predictions from the ANNs for flight service management are valid with respect to current industry standards for on-time arrival \citep{Hao2013}. 
    
\begin{table}[ht!] 
\begin{center}
\caption{Input features and corresponding values for a specific disrupted DAL-HOU flight} \label{tab:DAL-HOU_inputs}
\begin{tabular}{*3c}
    \hline
	    \textbf{Input Feature} &\textbf{ADM Phase}  & \textbf{Standardized Value}\\ \hline 
	    \textit{ADJST\_TURN\_MINS} & Tactical & -0.63548 \\ 
	    \textit{shiftper\_sched\_PB} & Operational & -0.07008 \\ 
	    \textit{shiftper\_sched\_GP} & Operational & 0.66334\\ 
	    \textit{DOT\_DELAY\_MINS} & Operational &  0.08585\\
	    \textit{shiftper\_actl\_PB}  & Strategic & 0.39709\\  
	    \textit{shiftper\_actl\_GP} &  Strategic & 1.38748\\  
	    \textit{actl\_enroute\_mins}  & Strategic & -0.49476 \\
	    \textit{ATC Hold at Origin (HD06)} & Operational & 1.12319\\
	    \textit{doy} & All & -0.01179\\
	    \textit{orig\_x\_dir} & All & -0.25736\\
	    \textit{orig\_y\_dir} & All & -0.75922\\
	    \textit{orig\_z\_dir} & All & -0.50746\\
	    \textit{dest\_x\_dir} & All & -0.12307\\
	    \textit{dest\_y\_dir} & All & -1.46375\\
	    \textit{dest\_z\_dir} & All & -1.25591\\
	    \textit{ONBD\_CT} & All & 0.46063\\
	    \textit{route\_dist} & All & -1.16903\\
	    \textit{route\_originator\_flag} & All & -0.36540\\
	   \hline
\end{tabular}
\end{center}
\end{table}
    
\subsection{Multi-Phase Results}
    
\begin{table}[ht!] 
\begin{center}
\caption{Target features and corresponding values for a specific disrupted DAL-HOU flight} \label{tab:DAL-HOU_target}
    \begin{tabular}{*3c}
        \hline
	    \textbf{Target Feature} &\textbf{Value Source}  & \textbf{Actual Value}\\ \hline
	    \textit{Actual A0} & Real Schedule Execution & 0 \\ 
	    \textit{Actual A14} & Real Schedule Execution &  0 \\
	    \textit{Scheduled Turnaround (mins)} & Real Schedule Execution & 35.00\\
	    \textbf{\textit{Actual Turnaround (mins)}} & Real Schedule Execution & \textbf{38.00}\\
	    \textit{Scheduled Block Time (mins)} & Real Schedule Execution & 65.00\\
	    \textbf{\textit{Actual Block Time (mins)}} & Real Schedule Execution & \textbf{104.00}\\
	    \textbf{\textit{Actual Tactical Delay (mins)}} & Real Schedule Execution & \textbf{3.00}\\
	    \textbf{\textit{Actual Strategic Delay (mins)}} & Real Schedule Execution & \textbf{39.00}\\
	    \textit{Predicted A0} & Operational ANN & 0 \\ 
	    \textit{Predicted A14} & Operational ANN & 0\\ 
	    \textit{Non-disrupted Turnaround (mins)}  & Tactical ANN & 30.03\\ 
	    \textit{Disrupted Turnaround (mins)} &  Tactical ANN & 38.37\\ 
	    \textit{Non-disrupted Block Time (mins)}  & Strategic ANN & 92.02 \\
	    \textit{Disrupted Block Time (mins)} & Strategic ANN & 93.61\\
	    \textbf{\textit{Estimated Turnaround (mins)}} & PTFM Inference & \textbf{38.37}\\
	    \textbf{\textit{Estimated Block Time (mins)}} & PTFM Inference & \textbf{93.61}\\
	    \textbf{\textit{Estimated Tactical Delay (mins)}} & PTFM Inference & \textbf{8.34}\\
	    \textbf{\textit{Estimated Strategic Delay (mins)}} & PTFM Inference & \textbf{1.59}\\
	    \hline
\end{tabular}
\end{center}
\end{table}

We now employ multi-phase inference of the set of ANNs for ADM to demonstrate the PTFM estimations of turnaround duration and block time. We utilize a specific test flight schedule from Dallas to Houston, delayed due to air traffic control hold for bad weather in Dallas.
    
Table~\ref{tab:DAL-HOU_inputs} reveals the inputs of respective ANNs for different phases of airline disruption and resource management for the disrupted (test) flight schedule from Dallas to Houston. Features in Table~\ref{tab:DAL-HOU_inputs} that serve as general inputs of all ANNs are determinate aleatoric features, which define the specificity (or uniqueness) of the test flight schedule under evaluation. Features in Table~\ref{tab:DAL-HOU_inputs} that serve as specific (additional) inputs of ANNs are epistemic features, which define the observability and peculiarity of the disruption resolution of a particular flight schedule during schedule execution.
    
Table~\ref{tab:DAL-HOU_target} shows the predicted values of target features from separate ANNs for ADM and actual values from the real world execution of the disrupted flight from Dallas to Houston, defined by the ANN (or PTFM) input information from Table~\ref{tab:DAL-HOU_inputs}. As evidenced in Table~\ref{tab:DAL-HOU_target}, the ANNs for operational ADM predict that the flight would not arrive at Houston as scheduled nor within $14 mins$ of the scheduled arrival time (i.e. binary indicators of 0), which is consistent with the real execution of the flight from Dallas to Houston. The ANNs for tactical ADM predict a turnaround duration of $30.03 mins$ and $38.37 mins$, repectively, for \textit{non-disrupted} and \textit{disrupted} flight capacity management of aircraft boarding at Dallas. In complement, the respective ANNs for strategic ADM predict a block time duration of $92.02 mins$ and $93.61 mins$ for \textit{non-disrupted} and \textit{disrupted} flight service management of aircraft operation from pushback at Dallas to gate-parking in Houston. Since the predicted $A0$ and $A14$ values sum to zero, the estimated turnaround and block time duration from the PTFM are the corresponding values of turnaround duration and block time retrieved from the \textit{disrupted} flight capacity and \textit{disrupted} flight service management ANNs. As such, the PTFM estimates of turnaround duration and block time are $38.37 mins$ and $93.61 mins$, respectively, for the disrupted flight from Dallas to Houston. 
    
The actual observed turnaround duration and block time for the disrupted DAL-HOU flight are $38 mins$ and $104 mins$ respectively. This indicates a 0.97\% difference between the PTFM estimate for turnaround duration and the actual turnaround duration of the weather-disrupted DAL-HOU flight. Similarly, there is a 10.52\% difference between the PTFM estimate for block time and the actual observed block time of the weather-disrupted flight from Dallas to Houston. The execution of the disrupted flight schedule from Dallas to Houston resulted in a tactical delay of $3 mins$ during turnaround at Dallas and a strategic delay (i.e. discretionary holdup by human specialists in the AOCC) of $39 mins$ from aircraft pushback at Dallas to aircraft parking at the arrival gate in Houston. On the other hand, the PTFM (i.e. system of ANNs) estimated a tactical delay of $8.34 mins$ and a strategic delay of $1.59 mins$ for this particular disrupted flight schedule from Dallas to Houston. As such, there is a 94.18\% difference between the tactical delay applied to resolve the disrupted DAL-HOU flight during schedule execution when compared to the tactical delay estimate from the PTFM. Similarly, the percentage difference between the strategic delay applied to resolve the disrupted flight during schedule execution and the strategic delay estimated by the PTFM is about 180\%. The PTFM-estimated values of block time and turnaround duration are comparable to the real-world flight schedule execution as these are deterministic processes that can be accurately modeled and estimated. In addition, multi-phase results revealed very small differences between the values of predicted block time and turnaround duration from the PTFM compared to the values observed from the real world execution of the flight. On the other hand, the PTFM estimates for tactical and strategic delay differ from the real world values as these are subjective values dependent on the discretion of a human specialist who decides how much delay to apply.   

\section{Conclusion} \label{conclusion}
Proactive and reactive disruption management (ADM) measures, for optimal flight schedule execution on day of operation, are representative of an airline's ability to efficiently manage it resources. This paper discussed and demonstrated the development and effectiveness of a modular system of artificial neural networks (ANNs), which represent a predictive transfer function model (PTFM) for expeditiously obtaining proper estimates of turnaround duration and block time duration during airline disruption management (ADM). The evaluation of the system of ANNs for multiple functional roles in the airline operations control center (AOCC), at a major U.S. airline, revealed that a characteristic PTFM for any functional role (i.e. intelligent agent) in the AOCC satisfies existing industry benchmarks for lateness during the execution and management of disrupted flight schedules. 

Single phase results revealed that an ANN topology consisting of a single hidden layer was sufficient in realizing optimal structures of the ANNs for all phases of ADM. This was evidenced by high performance of respective ANNs when analyzed with test data, such that ANNs for tactical, operational, and strategic ADM provided close comparisons in terms of pre-defined performance parameters for the different flight phases. However, multi-phase results revealed large differences between the values of predicted tactical and strategic delays from the PTFM compared to the real world values from flight schedule execution. These large differences arise due to the unexplained variance in discretionary holdup, and thus difficulty for the PTFM to predict how much tactical and strategic delay a human specialist will apply during ADM. 

While we present a novel modular approach to assemble a system of ANNs for ADM, the data used for this work was provided by a single U.S. airline that operates a point-to-point route network structure. As such, the PTFM created from using this data may not completely represent industry-wide disruption resolution propensities. Thus, a future research direction is to use additional data from other air transportation stakeholders (e.g. airports, unmanned aircraft systems (UAS) and UAS operations) and airlines that operate a hub and spoke route network. This will bolster the efficacy of the PTFM for industry-wide initiatives such as FAA collaborative decision making and emerging operations such as UAS operations in the national airspace. Furthermore, the PTFM can provide a predictive and prescriptive artificial intelligence (AI) component in the synthesis of AI and distributed ledger technology (i.e. Blockchain), thus enabling decentralized and distributed AI for ADM.        

\section*{Acknowledgement}
The authors would like to thank Blair Reeves, Chien Yu Chen, Kevin Wiecek, Jeff Agold, Dave Harrington, Rick Dalton, and Phil Beck, at Southwest Airlines Network Operations Control (SWA-NOC), for their expert inputs in abstracting the data used for this work. 

\section*{Conflict of Interest}
The authors have no conflict of interest to report. 

\bibliography{mendeley}

\end{document}